\pdfoutput=1
\documentclass[11pt]{article}

\usepackage{acl}

\usepackage{times}
\usepackage{latexsym}
\usepackage{booktabs}
\usepackage[normalem]{ulem}
\useunder{\uline}{\ul}{}
\usepackage[T1]{fontenc}
\usepackage[utf8]{inputenc}

\usepackage{microtype}

\usepackage{inconsolata}

\usepackage{amsmath}
\usepackage{graphicx}
\usepackage{booktabs}
\usepackage{subcaption}
\usepackage{tcolorbox}
\usepackage{multirow}
\usepackage{xcolor}
\usepackage{listings}

\lstdefinestyle{mystyle}{
    basicstyle=\footnotesize\ttfamily,
    backgroundcolor=\color{pink!20},  
    breaklines=true,                 
    breakatwhitespace=true,            
}

\usepackage{enumitem}
\newcommand*{\affmark}[1][*]{\textsuperscript{#1}}
\newcommand*{\email}[1]{\texttt{#1}}
\setlist[itemize]{itemsep=0mm}

\title{Cooperation on the Fly: Exploring Language Agents for Ad Hoc Teamwork in the Avalon Game}

\author{Zijing Shi\affmark[1], Meng Fang\affmark[2], Shunfeng Zheng\affmark[1], Shilong Deng\affmark[2], Ling Chen\affmark[1], Yali Du\affmark[3] \\
\affmark[1] AAII, University of Technology Sydney, Australia\\ 
\email{\{zijing.shi, shunfeng.zheng\}@student.uts.edu.au,
ling.chen@uts.edu.au} \\
\affmark[2] University of Liverpool, United Kingdom \\
\email{\{Meng.Fang, shilong.deng\}@liverpool.ac.uk} \\
\affmark[3] King’s College London, United Kingdom \\
\email{yali.du@kcl.ac.uk}
}

\begin{document}
\maketitle
\begin{abstract}
Multi-agent collaboration with Large Language Models (LLMs) demonstrates proficiency in basic tasks, yet its efficiency in more complex scenarios remains unexplored.
In gaming environments, these agents often face situations without established coordination protocols, requiring them to make intelligent inferences about teammates from limited data. 
This problem motivates the area of ad hoc teamwork, in which an agent may potentially cooperate with a variety of teammates to achieve a shared goal. 
Our study focuses on the ad hoc teamwork problem where the agent operates in an environment driven by natural language.
Our findings reveal the potential of LLM agents in team collaboration, highlighting issues related to hallucinations in communication.
To address this issue, we develop CodeAct, a general agent that equips LLM with enhanced memory and code-driven reasoning, enabling the repurposing of partial information for rapid adaptation to new teammates.

% These insights are crucial for employing LLM agents in real-world collaborative scenarios.
\end{abstract}

\section{Introduction}

% Recent advancements in LLM agent 
Large Language Models (LLMs) have exhibited impressive abilities in reasoning and generalization, showing their potential in building autonomous agents and driving many advancements \cite{kojima2022large, xi2023rise}.
Recent research involving multiple LLM agents working together has shown success in task execution \cite{wang2023unleashing, wu2023autogen, dong2023self}.
However, these studies often employ explicit coordination protocols for fully cooperative tasks, assuming full information of the teammate and the task. 
While some research has explored simulating LLM agents in more open-ended settings, these agents remain limited by pre-established strategies and behaviors \cite{park2023generative}.

% Introducing AHT
In dynamic environments, maintaining consistent coordination protocols among agents is a challenge. Such contexts demand an agent capable of swift adaptation and on-the-fly cooperation. 
Without pre-established team strategies, an agent, often referred to as the learner, must observe its teammates' behaviors and adapt effectively to achieve their shared goals. This challenge is known as the Ad Hoc Teamwork (AHT) problem \cite{stone2010ad, macke2021expected}. 

% AHT with LLM agents
A significant challenge for the learner is to infer the role of their teammates. 
Traditional methods rely on pre-programmed rules or data-driven algorithms trained on historical data \cite{mirsky2022survey}. In contrast, LLM agents can directly communicate with their teammates in natural language. However, the communication dynamics within the environment may exhibit conflicting and competitive elements. Moreover, due to the inherent traits of LLMs, teammates may provide inaccurate or misleading information.

% Introducing AvalonPlay benchmark
In this study, we explore the AHT abilities of LLM agents, utilizing the multi-agent game \textit{The Resistance: Avalon} to simulate AHT environments.
We introduce the AvalonPlay benchmark, which challenges agents with hidden roles to participate in strategic gameplay across multiple rounds. 
In each round, a learner is tasked with selecting teammates without prior strategies and full information.
Our research integrates LLM agents into AvalonPlay to evaluate their performance in AHT. Rather than developing coordination protocols for the entire team, we prioritize designing agents adept at cooperating in environments without such explicit protocols.

\begin{figure*}[ht!]
  \centering
  \includegraphics[width=\linewidth]{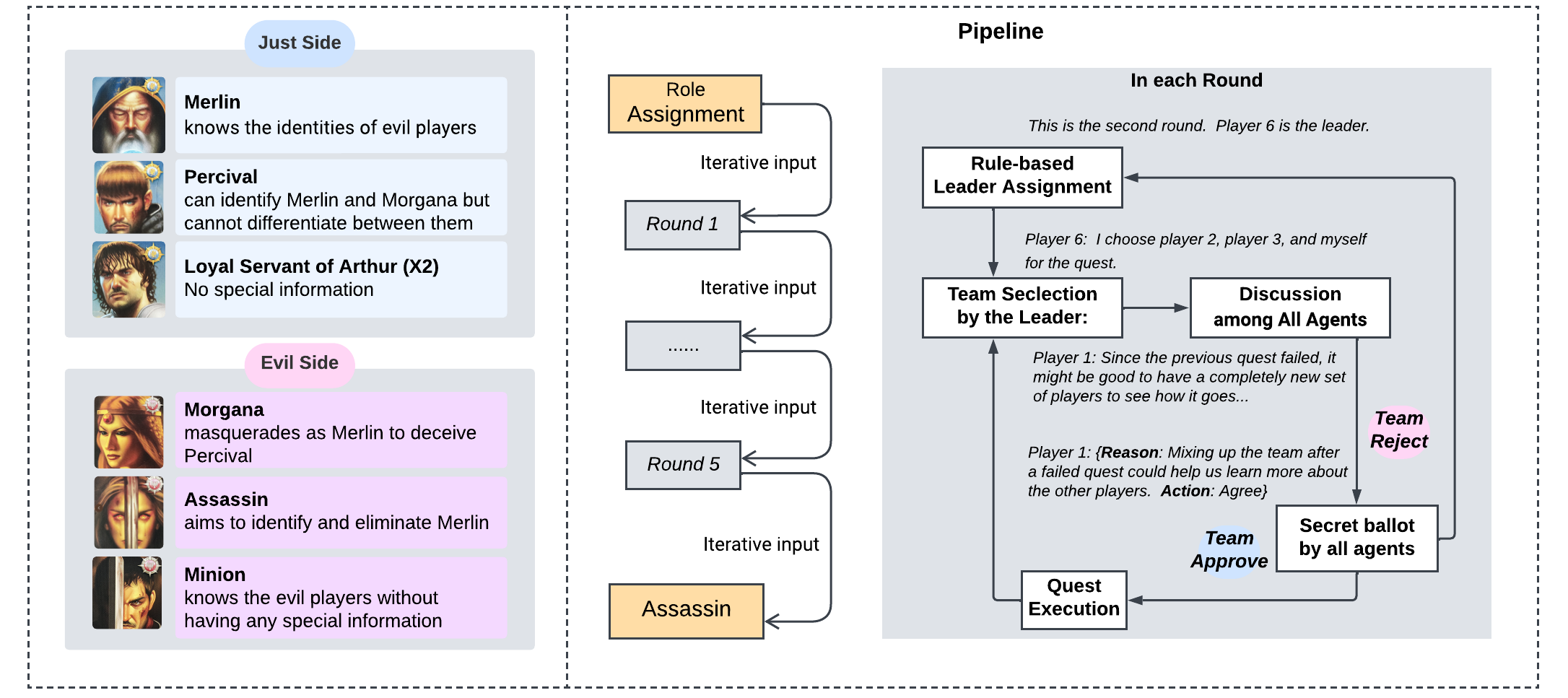}
  \caption{A flowchart of the AvalonPlay benchmark showing team sides and roles on the left and a detailed round pipeline on the right. Each round includes four stages: leader assignment, team selection, discussion and voting, quest execution.}
  \label{fig:avalonflowchart}
\end{figure*}

% Methods
A communication channel is designed within the AvalonPlay benchmark to enhance collaboration. We observe that LLM agents may sometimes produce hallucinations and forget early information. To address this, we introduce CodeAct, a general agent leveraging the advanced reasoning capabilities of LLMs, focusing on reasoning about various potential teammate types. 
We begin with a memory retrieval system that collects information and utilize the knowledge gained from these interactions. Subsequently, we incorporate code-driven reasoning with action to effectively refer the roles of new teammates. With the teammate role information, our agent is well-equipped to effectively cooperate with its teammates.

% Contribution
In summary, our contributions are as follows: 
\begin{itemize}
    \item We explore the AHT capabilities of LLM agents, utilizing the AvalonPlay benchmark with its multi-round collaborative tasks as a testing ground for these abilities.
    
    \item We develop CodeAct, an LLM agent utilizing a code-driven prompting approach, aimed at boosting the agent's efficiency in AHT tasks. 
    
    \item We conduct comprehensive experiments within the AvalonPlay benchmark, demonstrating the effectiveness of our proposed agent.
\end{itemize}

\section{The AvalonPlay Benchmark}

We introduce the AvalonPlay benchmark to explore the performance of LLM agents on AHT. This benchmark is a language-based, multi-agent platform comprised of multi-round tasks. During the gameplay, agents have limited knowledge about each other's roles. In each round, a team leader, also regarded as the learner, is selected to deduce the roles and intentions of other agents. The leader must adapt their team's strategies dynamically as the game evolves.  
This section details the key elements of the proposed benchmark.

\subsection{Teammate Role}

We utilize a setup involving seven players indexed by $j$, each controlled by an LLM agent. In each game, agents are randomly assigned one of six distinct roles, dividing them into two factions: the just side (including \texttt{Merlin, Percival}, and two \texttt{Loyal Servants of Arthur}) and the evil side (including \texttt{Morgana, Minion}, and \texttt{Assassin}).

Each role possesses unique abilities, as shown on the left of Figure \ref{fig:avalonflowchart}. 
On the just side, \texttt{Merlin} is aware of the factions but not the specific roles, while \texttt{Percival} can recognize \texttt{Merlin} and \texttt{Morgana} but is unable to distinguish between them. The \texttt{Loyal Servants of Arthur} lack any special knowledge. On the evil side, in contrast, all roles can recognize their teammates.
This setup puts the just side at a disadvantage due to their limited information. Our study focuses on scenarios where the just side takes the lead role, aiming to enhance the leader's ability to deduce the various roles of agents.

\begin{figure*}[t!]
  \centering
  \includegraphics[width=\linewidth]{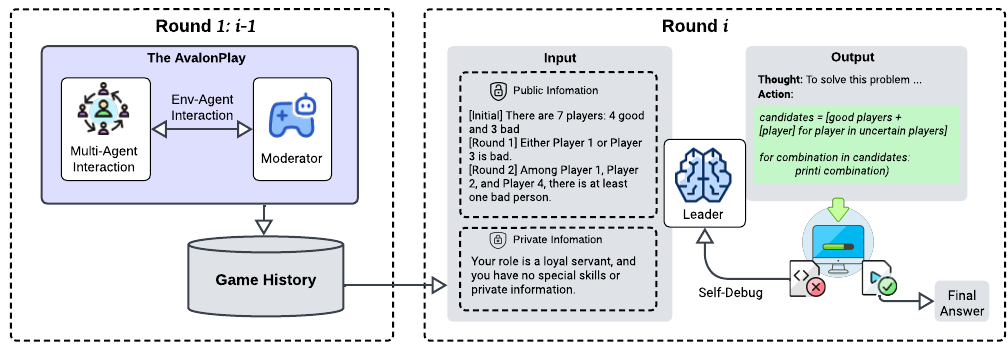}
  \caption{An overview of the proposed CodeAct agent as the leader during team selection. We begins by establishing a memory retrieval system that distills information from past interactions, enabling the agent to access relevant information. Then, we integrate code-driven reasoning with action to determine teammate roles effectively. Finally, we employ an interpreter to execute the generated code, equipping the agent with self-debug capabilities.}
  \label{fig:agent}
\end{figure*}

\subsection{Pipeline}

In AvalonPlay, a single game is composed of five quest rounds indexed by $i$. Each round, defined as a full quest cycle, is divided into the following four phases, as shown in the right of Figure \ref{fig:avalonflowchart}.

\begin{itemize}[itemsep=0mm, leftmargin=10pt]
     \item \textbf{Leader Assignment:} Each round begins with the selection of a leader who nominates teammates for the quest. The first round's leader is randomly chosen, and in subsequent rounds, leader rotates sequentially to the next player. This approach ensures diverse leader roles while incorporating randomness.
    \item \textbf{Team Selection:} In this phase, a leader nominates $n_i$ teammates, with the team size $n_i$ varying per round. Adhering to the game rule, the team sizes are: $n_1=2, n_2=3, n_3=3, n_4=4, n_5=4$. We align the leader's response into corresponding actions. For instance, the response ``\texttt{I choose myself and player 2 for the quest team}'' is mapped to the specific selection of teammates.
    \item \textbf{Discussion and Voting:} We design a communication protocol that enables the agent to acquire information while accounting for the potential of misinformation. Post team selection, agents participate in discussions to assess the proposed team. Following the discussion, each agent secretly votes $\mathbf{1}_{j}$ indicates agent $j$'s approval of the current nominated team. 
    To enhance agents' decision-making, they are required to generate a JSON containing their reasoning and vote. 
    A team is approved if the total votes reaches a majority (i.e., $\sum_{j=1}^{7} \mathbf{1}_{j} \geq 4$). Otherwise, the team selection is repeated with a new leader. A team is automatically formed if approval is not achieved after five consecutive attempts.
    \item \textbf{Quest Execution:} For an approved team, teammates from the just side must vote to approve, while teammates from the evil side must vote against. Let $Q_i$ denote the total votes of evil members. The quest is sabotaged if $Q_i \geq 1$ for rounds $i \in \{1,2,3\}$, or $Q_i \geq 2$ for rounds $i \in \{4,5\}$. In other cases, it is completed successfully.
\end{itemize}

We adhere to game rule by including an assassination phase in the Avalonplay benchmark, where the \texttt{Assassin} identify \texttt{Merlin} after all rounds. However, our study aims to assess the teamwork of LLM agents, not the accuracy of assassinations, though this feature is included for prospective future research.

\subsection{Observation Understanding}

The AvalonPlay benchmark includes a moderator responsible for overseeing the game's progress. This involves providing action descriptions to agents and recording events that occur during gameplay. The moderator's role is non-interactive with the LLMs but adheres to predefined procedures. The information generated by the moderator is a vital part of the agents' observation.

We develop structured prompts to assist
LLM agents in comprehending both the game’s rules and its current state. 
% We first develop a global prompt to assist LLM agents in understanding Avalon's rules and their assigned roles. The global prompt is consistently fed to the agents during each interaction round, as shown in Table \ref{tab:global_prompt}. We also incorporate the role identification and prior messages as real-time inputs, effectively reflecting the game's current state. 
In general, an LLM agent's observation is divided into four key variables to navigate the game:
\begin{itemize}[itemsep=0mm, leftmargin=10pt]
    \item \textbf{Game Rule:} We begin with a basic understanding of the game (e.g., ``\texttt{You are playing the Avalon game with other players...}'').
    \item \textbf{Role Assignment:} For each player, we randomly assign their roles and create a unique role assignment prompt (e.g., \texttt{Merlin} is informed: ``\texttt{You are Merlin...Your objective is to ensure the success of the quests as much as possible}'').
    \item \textbf{Role Identification:} Roles with special abilities get extra information (e.g., \texttt{Merlin} receives a moderator's note: ``\texttt{Players 1, 2, and 6 belong to the evil side}'').
    \item \textbf{Game History:} This includes the game trajectory from previous rounds, as well as earlier time steps of the current round.
    \item \textbf{Action Description:} This includes a description of the next possible actions for the agent, provided by the moderator (e.g., Player 2 receives a moderator's note: ``\texttt{Now it's your turn, player 2. Please discuss whether to agree with the nominated team}'').
\end{itemize}

The game rule, role assignment, and role identification collectively constitute the global prompt, consistently employed at each time step. Game history and action description serve as real-time prompt, effectively capturing the game's present state. For additional details, please refer to Appendix \ref{appendix_prompt}.

\section{Methodology}

\subsection{Overview}

Reasoning about various potential teammate types is a key issue in AHT research. To improve the agent's reasoning abilities, we introduce the CodeAct agent, which utilizes the code-based reasoning capabilities of LLMs. Our CodeAct agent does not focus on creating prompts or demonstrations for specific tasks. Instead, it emphasizes a general strategy that transforms complex semantic tasks into flexible code structures.

Our approach unfolds in three stages, as shown in Figure \ref{fig:agent}. Firstly, we establish a memory retrieval system to extract information from previous interactions. 
Then, we integrate the code-driven reasoning with action to reason teammate roles. Lastly, we employ an interpreter to execute the code and equip the agent with the self-debug capability.

\subsection{Memory Retrieval}
% Within the Avalonplay benchmark, a designated moderator documents the game's progression and factual occurrences. 
During game playing, we implement global memory for each LLM agent to prevent the over-extension of token inputs. 
Particularly during the team selection phase, the team leader is equipped with a customized memory module, strategically designed to aid in deducing the roles of its teammates.

\paragraph{Global Memory}
The global memory for each agent comprises identified key information along with messages from the most recent $k$ steps. This key information includes: 1) Notes from the game moderator; 2) Contributions from agents during discussions, which reveal details of personal identities or deduce the identities of others.

\paragraph{Leader-Specific Memory}
LLM agents often base their reasoning on the latest information and frequently experience hallucinations, as detailed in Section \ref{quanti}. In response, for each game round's team selection stage, we systematically retrieve factual data gathered from earlier interactions to aid the current leader. 
This includes:
1) Private Information, derived from unique skills tied to an agent’s role, like Percival’s insight: ``\texttt{Among players 1 and 4, one is good and the other evil, but you can't discern who is who.}''
2) Public Information, pertaining to historical game outcomes and events, accessible to all agents, such as: ``\texttt{In the initial round, players 1 and 2 were selected for the quest, which ended in failure.}''

\lstset{style=mystyle}
\begin{figure}[ht]
\begin{lstlisting}
You are engaged in a game with 7 players, each potentially categorized as 'good' or 'bad'. There are 4 good players and 3 bad players among them. 
You have the fllowing information.
{PRIVATE_INFORMATION}
{PUBLIC_INFORMATION}
Please generate a code program that selects {self._current_team_number} players most likely to be good person. Please print the final 4 players.'''})

\end{lstlisting}
\caption{An example of the CodeAct agent's prompt.}
\label{fig:prompt}
\end{figure}

\lstset{style=mystyle}
\begin{figure}[ht]
\begin{lstlisting}
Action: Python_REPL
Action Output: 
players = [None] * 7  # Initialize the players list with None
# Apply Round 2 information
players[1] = True  # Player 2 is good
players[3] = True  # Player 4 is good
players[6] = True  # Player 7 is good

# Apply Round 3 information
# Since Player 2 and Player 4 are good, Player 1 must be bad
players[0] = False  # Player 1 is bad

# Apply Round 4 information
# Since Player 2 and Player 7 are good, and we know Player 1 is bad, Player 3 must be good
players[2] = True  # Player 3 is good
# This leaves Player 5 as the only possible bad player from this group
players[4] = False  # Player 5 is bad

# We already have 2 bad players (Player 1 and Player 5), so Player 6 must be good
players[5] = True  # Player 6 is good

# Print the final 4 players most likely to be good
good_players = [index + 1 for index, is_good in enumerate(players) if is_good]
print(good_players)

Observation: [2, 3, 4, 6, 7]
Final Answer: The teammates are Players 2, 3, 4, and 6.
\end{lstlisting}
\caption{An example of the CodeAct agent's output.}
\label{fig:output}
\end{figure}

%\subsection{Code-driven Reasoning with ReACT}
\subsection{Code-driven Reasoning with Action}

In the team selection phase, we utilize a customized memory system, allowing the current leader to deduce the roles of its teammates. Drawing inspiration from ReAct \cite{yao2022react}, our agent is designed for interactive generation of both thoughts and actions through a program in a zero-shot manner. We utilize a program that processes useful information by reasoning over the agent's current private and public data to support future actions. As illustrated in Figure \ref{fig:prompt}, we further enhance this process by enabling the leader to generate reasoning substeps in a code-like format.  
In this study, we focus on Python, but our approach is applicable to any programming language. 
Figure \ref{fig:output} provide an excerpt generated by our CodeAct, with more details available in Appendix \ref{appendix_output}.

Our code-driven approach, on the one hand, leverages advantages from semantic reasoning that involve intermediate steps. On the other hand, it leverages code-based reasoning, enabling reliable inference through the formal yet expressive code structure and the powerful computational tools it offers.

\begin{figure*}[htp]
    \centering
    \begin{subfigure}[b]{0.3\textwidth}
        \centering
        \includegraphics[width=\textwidth]{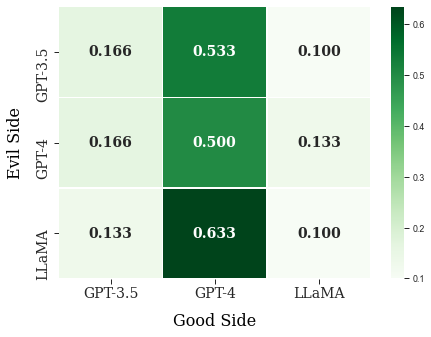}
        \caption{Game Win Rate by Quest}
        \label{fig2:sub1}
    \end{subfigure}
    \hfill
    \begin{subfigure}[b]{0.3\textwidth}
        \centering
        \includegraphics[width=\textwidth]{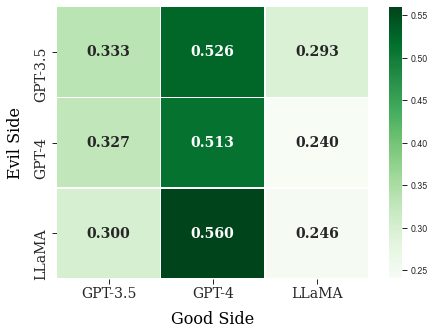}
        \caption{Total Quest Win Rate}
        \label{fig2:sub2}
    \end{subfigure}
    \hfill
    \begin{subfigure}[b]{0.3\textwidth}
        \centering
        \includegraphics[width=\textwidth]{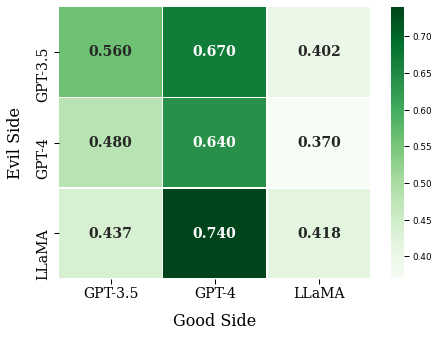}
        \caption{Team Selection Accuracy}
        \label{fig2:sub3}
    \end{subfigure}
    \caption{The results of different models acting as the good and evil sides, playing games against each other. Under each setting, 30 games were conducted, totaling 150 quests.}
    \label{fig:winrate}
\end{figure*}

\begin{figure*}[htp]
    \centering
    \begin{subfigure}[b]{0.3\textwidth}
        \centering
        \includegraphics[width=0.9\textwidth]{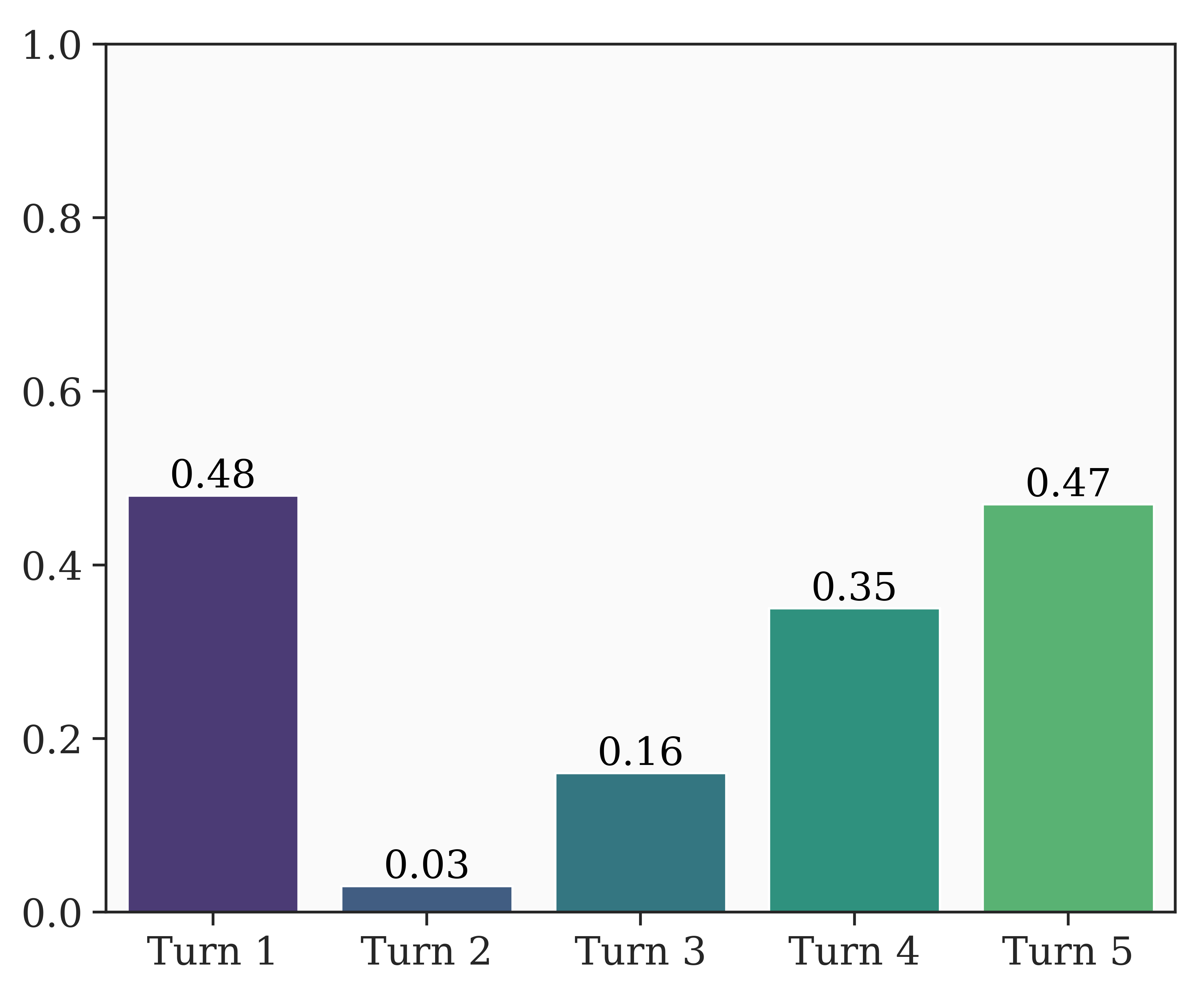}
        \caption{GPT-3.5 as Good Side}
        \label{fig3:sub1}
    \end{subfigure}
    \hfill
    \begin{subfigure}[b]{0.3\textwidth}
        \centering
        \includegraphics[width=0.9\textwidth]{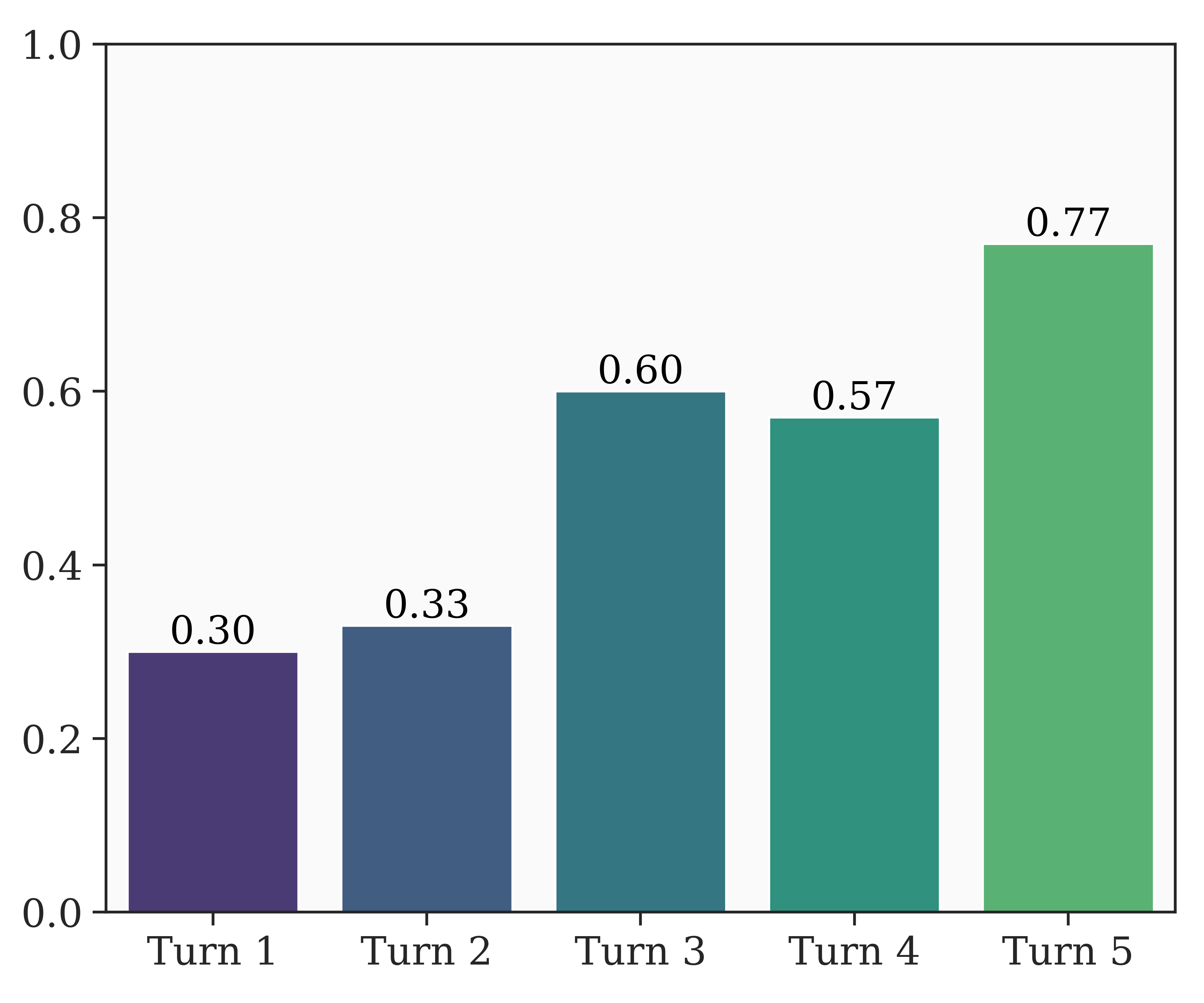}
        \caption{GPT-4 as Good Side}
        \label{fig3:sub2}
    \end{subfigure}
    \hfill
    \begin{subfigure}[b]{0.3\textwidth}
        \centering
        \includegraphics[width=0.9\textwidth]{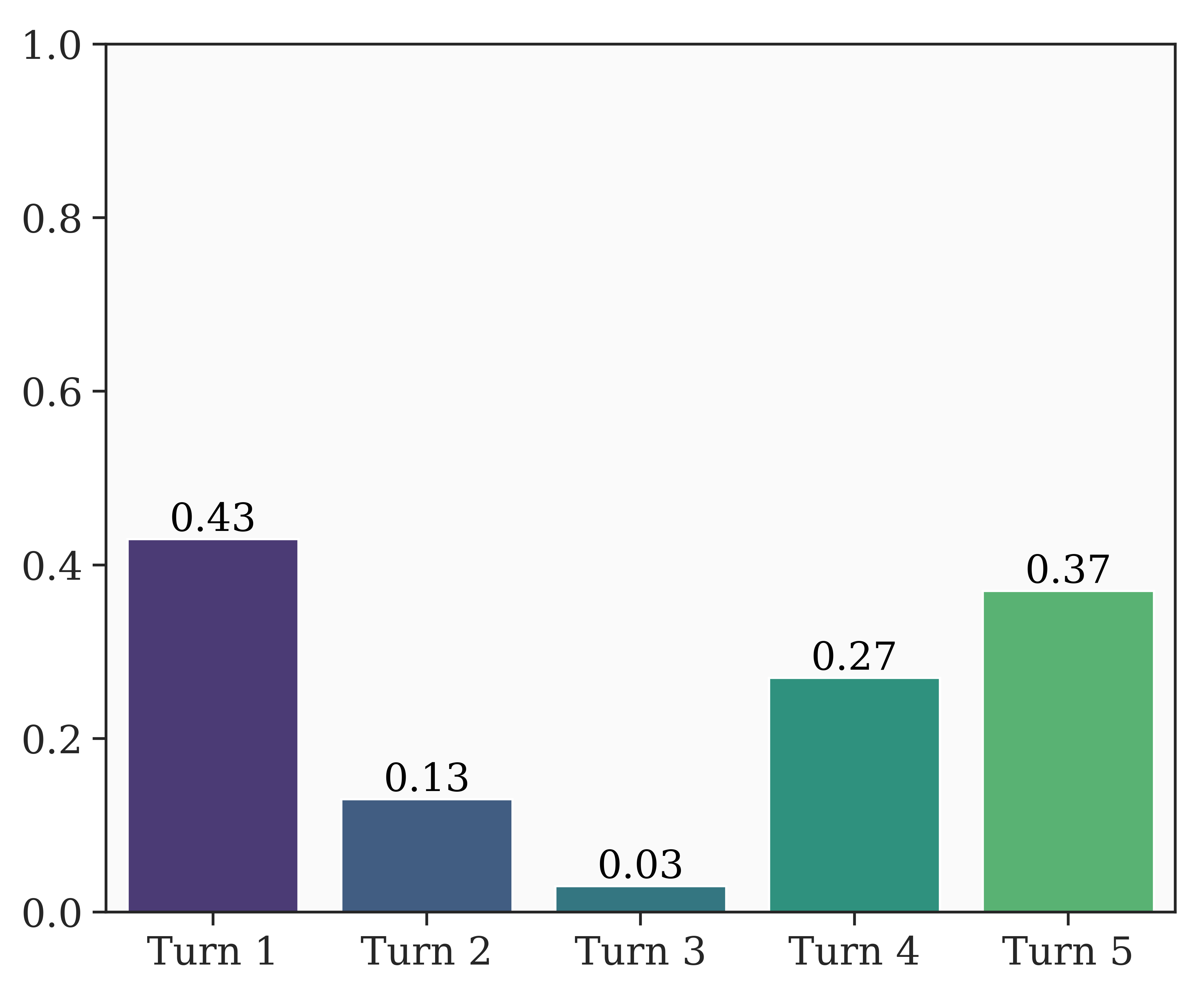}
        \caption{LLaMA2 as Good Side}
        \label{fig3:sub3}
    \end{subfigure}
    \caption{The turn-by-turn quest win rate results when different models assume the role of the good side. Each model, as the good side, played 90 games, resulting in 450 quests evenly distributed across 5 turns.}
    \label{fig:round}
\end{figure*}

\subsection{Code Execution with Self-Debug}
We equip the leader with the reflection mechanism \cite{shinn2023reflexion, chen2023teaching}, allowing them to refine their programs. The code-based action, once formulated, is then processed by a code interpreter. In this study, we equip the agent with a Read-Eval-Print-Loop Python interpreter (i.e., \texttt{Python\_REPL}). 
If the code executes successfully, the agent's decision is informed by the resulting program output. Conversely, if execution falters, the agent will revise the code, taking into account the issue encountered and previous outputs to enhance its generation.

\section{Experiments}

Our experiments aim to explore the following questions:
1) The AHT capabilities of agents employing different LLMs as backend models; 
2) The effect of a natural language-based communication protocol on the AHT capabilities of LLM agents; 
3) The impact of diverse prompting strategies on the AHT capabilities of LLM agents. We compare the proposed CodeAct agent with other semantic-based reasoning approaches.

\subsection{Baselines}

To tackle question 1, we conduct a preliminary evaluation of three advanced LLMs within the multi-agent AvalonPlay benchmark. These models include GPT-3.5, GPT-4, and LLaMA 2. Specifically, we utilize the versions \texttt{gpt-3.5-turbo-0613}, \texttt{gpt-4-1106-preview}, and \texttt{Llama2-chat-13B} respectively. These LLM agents, equipped with different backend LLMs, are deployed on both the good and evil sides to compete against one another.

To address question 2, we compare scenarios in AvalonPlay both with and without communication protocols. In scenarios lacking communication protocols, all players proceed to a secret vote immediately following the team leader's selection of the team.

To tackle question 3, we focus on the team selection by the leader, compare semantic reasoning approaches (i.e., Chain of Thought \cite{wei2022chain} and ReAct \cite{yao2022react}) with our proposed CodeAct agent. 
% Additionally, we perform detailed ablation experiments to gain deeper insights into the influence of various components. 

\subsection{Metrics}

We design the following three metrics to measure an LLM agent's performance in the AvalonPlay benchmark.
\begin{itemize}[itemsep=0mm, leftmargin=10pt]
    \item \textbf{Game Win Rate by Quest}  (\texttt{Game Win}), which refers to the percentage of games where the good side wins by completing three out of five quests.
    \item \textbf{Total Quest Win Rate} (\texttt{Quest Win}), which refers to the percentage of quests where the good side wins.
    \item \textbf{Team Selection Accuracy} (\texttt{Team Acc}), which refers to the percentage of times the just side, when in the role of leader, selects the team for a successful quest.
\end{itemize}

\subsection{Implementation Details}

% Our analysis focus on the performance of different agents in various rounds. To ensure a more comprehensive analysis, we carry out each game for 5 rounds, regardless of whether the outcome could be determined in earlier rounds. 
% In each setting, we conduct 30 games and finally obtain 150 quest cycles.

For all LLM agent, we set the sampling temperature as 0.7 for text generation and 0 for code generation. We set the window size $k$ to be 15 for the global memory retrieval.
During gameplay, all roles are randomly assigned to each player to avoid biasing the data towards certain players based on the turn order.

\subsection{Main Results}
\paragraph{Question 1: Model Performance.} 

Figure \ref{fig:winrate} displays the performance of agents with three different backend LLMs, playing on different sides. Figure \ref{fig:round} shows the quest success rates of these LLM agents as the good side across different rounds. Our findings indicate that 1) More advanced LLMs achieve higher win rates both in games and in overall tasks when playing as the good side, with GPT-4 agents exhibiting the best performance. 2) Notably, GPT-4 agents demonstrate higher quest success rates in later rounds compared to earlier ones, indicating their ability to reason from observations. In contrast, GPT-3.5 and LLaMA agents do not show significant progress across rounds. This suggests that the GPT-4 agent is more adept at learning and adapting, leading us to utilize the GPT-4 model in subsequent experiments.

\begin{table*}[ht!]
\resizebox{1\textwidth}{!}{%
\begin{tabular}{@{}lp{20cm}@{}}
\toprule
% \multicolumn{2}{c}{\textbf{The 1st Round Team Selection and Discussion}} \\ 
% Player 5: & I choose player 1 and player 4  for the quest. \textcolor{orange}{(Team selection by the current leader)}\\ 
% Player 1: & I agree with this quest team. Let's go forward and succeed in this quest. \\
% Player 2: & I think it's a good start. Let's   see how the quest goes with this team. I agree with it. \\
% Player 3: & I agree with the team selection   for this quest. Let's trust player 5's judgment and work towards a successful   quest. \\
% Player 4: & I also agree with this quest team. Let's aim for success! \\
% Player 5: & We should give it a try and see how the quest unfolds. I agree with this team. \\
% Player 6: & It's the first round, and we don't have much information yet, but I think it's important to be cautious. I would prefer a more diverse selection, perhaps including some players from the latter half of the roster to balance things out. I disagree with the current team selection for the quest. \\
% Player 7: & As a Loyal Servant of Arthur, I understand the importance of being cautious, especially in the early rounds where we have limited information. However, it is also vital to establish trust and gather information from the first quest. I agree with the team selection for now, as it will allow us to observe the outcomes and make more informed decisions in subsequent rounds.  \\ \midrule
\multicolumn{2}{c}{\textbf{The 5th Round Team Selection and Discussion}} \\ 
Moderator: & Player 3, as the team leader, please select 4 team members to undertake the quest. You may also choose yourself.\\
Player 3: & I choose player 2, player 3,  player 6, and player 7 for the quest. \textcolor{orange}{(Team selection by the current leader)} \\ 
Moderator: & The selected team by the leader include ['player 2', 'player 3', 'player 6', 'player 7']. Now everyone discuss if you agree the team. \\
Player 1: & This team includes players who have been part of successful quests before. It's a logical choice to maintain   a strong team dynamic. I agree with the team selection. \\
Player 2: & Player 3, player 6, and player 7 were on the last successful quest with me, and I believe this consistency will help us ensure another victory. I'm in favor of this team. \textcolor{orange}{(In the previous round, players 2, 3, 6, and 7 executed the quest together and succeeded. Player 2 effectively used this information for reasoning)} \\ 
Player 3: & I've chosen players who have either been part of successful quests or have not shown any suspicious   behavior so far. It's important to maintain a strong team dynamic, and I   believe this selection will lead us to victory. \\
Player 4: & The team seems solid with  previous successful quest members. Including player 6 could either solidify   their trustworthiness or provide us with more information. I agree to   proceed. \\
Player 5: & The inclusion of player 6, who hasn't been on a quest recently, along with proven successful quest members,   provides a chance to validate their loyalty. We should be wary, but I am inclined to give this team a chance. \textcolor{orange}{(Player 5, however, produced hallucination, with agent's statement being inconsistent with the facts shared by everyone)}\\
Player 6: & The team includes players who have consistently been part of successful quests, and as a trusted member   myself, I feel confident we can achieve success once more. I agree with this selection and am prepared to contribute to the quest's success. \\
Player 7: & The team selected by player 3 includes myself and other players who have been part of successful quests, which instills confidence in this group's potential for success. Player 6's inclusion is a variable, but it is essential to evaluate all players. Based on the track record of success we've had, I agree with this team selection and am optimistic about our prospects. \\ \bottomrule
\end{tabular}
}
\caption{An example of the GPT-4 agent's participation in discussions during gameplay. In the fifth round, the agent's response is based on previous observations and demonstrates some reasoning ability.
However, the agent utilizes only limited information and is prone to experiencing hallucinations.}
\label{tab:casestudy}
\end{table*}

\paragraph{Question 2: Communication Channel.}
Table \ref{table:communication} compares the GPT-4 agent's performance in self-playing, both with and without communication. The introduction of communication does not significantly enhance the game performance of LLM agents. This is possibly due to the fact that, besides the communication protocol, LLM agents also have access to observe and reason based on the outcomes of quests in previous rounds. These findings indicate that communication does not always lead to improved collaborative efficiency in LLM agents. We further explore this phenomenon in Section \ref{quanti}.

\begin{table}[t]
\centering
\caption{Results from the GPT-4 agent self-playing in the AvalonPlay benchmark, both with and without communication. Under each setting, 30 games were conducted, totaling 150 quests.}
\label{table:communication}
\resizebox{\columnwidth}{!}{%
\begin{tabular}{@{}c|ccc@{}}
\toprule
\textbf{Agent/ Metric}       & \textbf{Game Win} & \textbf{Quest Win} & \textbf{Team Acc} \\ \midrule
\textbf{Base}            & 0.500    & 0.513       & 0.640    \\
\textbf{Base W/O Communication} & 0.466    & 0.480     & 0.693\\ 

\bottomrule
\end{tabular}
}
\end{table}

\paragraph{Question 3: Code-driven Reasoning.}
We campare the proposed CodeAct with two semantic reasoning methods: Chain of Thought (CoT) and ReAct. Table \ref{table:code} shows that our proposed CodeAgent achieves the best performance across three metrics, with a team selection accuracy of 0.830, demonstrating the superiority of code-based reasoning.

\begin{table}[t]
\centering
\caption{Results of the GPT-4 agent self-playing in the AvalonPlay benchmark using different prompting strageties. Under each setting, 30 games were conducted, totaling 150 quests.}
\label{table:code}
\resizebox{\columnwidth}{!}{%
\begin{tabular}{@{}c|ccc@{}}
\toprule
\textbf{Agent/ Metric}       & \textbf{Game Win} & \textbf{Quest Win} & \textbf{Team Acc} \\ \midrule
\textbf{Base with CoT}           & 0.533    & 0.547       & 0.707    \\
\textbf{Base with ReAct} & 0.500    & 0.526     & 0.634\\ 
\textbf{CodeAct}  & 0.633    & 0.593     & 0.830\\ 
\bottomrule
\end{tabular}
}
\end{table}

% \subsection{Ablations}

% To validate the necessity of each component in our
% approach, we conducted a detailed ablation study. We remove each of the components in the pipeline of our method and empirically discuss how it will influence the model outputs.

\subsection{Quantitative Analysis\label{quanti}}

In our preliminary analysis of the experimental logs targeting question 1, we observe that employing LLM agents in multi-turn tasks often results in memory forgetting. In cooperative tasks involving communication, these agents tend to exhibit cascading hallucinations and logical inconsistencies. We present Table \ref{tab:casestudy} for a more detailed illustration.

\paragraph{Forgetting Early Information.} Players typically focus on information from the last round, failing to integrate key information from the early stages of the game into their later decisions and strategies.

\paragraph{Hallucination Generation.} In the game, statements that contradict a player's actual role are not classified as hallucinations, as deception is a common strategy. Similarly, fabricating scenarios to falsely accuse others is also not considered hallucinations, as this mirrors the strategies of human players. We define inconsistencies within a response and counterfactual content in an iteration as hallucinations. 
When GPT-3.5 and LLaMa 2 models serve as agents, they tend to produce responses inconsistent with the prompt. A notable example is their role as team leaders, where they often select a number of teammates that diverges from the required amount. Additionally, the final votes they generate may not correspond with the reasoning process outlined in the previous step.
This problem is less pronounced with the GPT-4 agent during gameplay. However, all models, including GPT-4, are prone to generating responses that contradict established facts, such as modifying or denying events from previous game rounds. Table \ref{tab:casestudy} provides an example of GPT-4 agent’s participation in discussions. Player 5's statements are inconsistent with the facts. As player 2 points out, playe 6 had already successfully participated in the quest in the previous round.

This suggests that the communication protocol does not necessarily offer optimal solutions. The most important aspect is the effective verification of the factuality and correctness of communications. This distinguishes it from prior AHT research with agents trained from historical data \cite{mirsky2020penny}. 
In our study, we address this by utilizing memory retrieval to extract factual information. Additionally, we enhance the agent's reasoning abilities for different types of teammates by integrating code-based reasoning with actions.

\section{Related Work}

\paragraph{LLMs and Agents.} 

The advent of generative agents for interactive tasks marks a significant paradigm shift. Unlike traditional methods that relied on Deep Reinforcement Learning (DRL) \cite{shi2022stay,shi2023self,xu2020deep,xu-etal-2021-generalization-text,xu2022perceiving}, which confined agents to constrained settings, the emergence of LLMs has notably broadened the scope of possibilities \cite{openai2023gpt4}.
Efforts are increasingly focused on leveraging LLMs for complex tasks like planning and reasoning \cite{wang2023unleashing, wu2023autogen, dong2023self}. Innovative prompting methods, including ReACT \cite{yao2022react}, Chain of Thought (CoT) \cite{wei2022chain}, and Tree of Thought (ToT) \cite{yao2023tree}, significantly enhance the planning and reasoning capabilities of LLMs.
Some research integrates complex human cognitive functions into LLM agents, such as memory \cite{wang2023augmenting}, reflection \cite{shinn2023reflexion}, and tool usage \cite{huang2023metatool}. 

\paragraph{Multi-agent Interaction.}
Recent research has explored that utilizing the interactions among multiple LLM agents to achieve several goals \cite{chen2023agentverse, wu2023autogen}. These studies focus on two main areas.
First, it involves assigning distinct roles to agents to enhance their task performance \cite{wang2023survey}. Notable applications include using multiple agents in debates to counter self-reflection biases \cite{liang2023encouraging, chan2023chateval}, and diversifying agent roles for complex projects like software development \cite{qian2023communicative, hong2023metagpt}.
Second, researchers develop multiple LLM agents in sandbox environments to mimic real human behavior \cite{park2023generative, zhou2023far}. Some of these investigate the capabilities that emerge in multi-agent settings, such as deception and leadership \cite{xu2023exploring}.

\paragraph{Ad Hoc Teamwork.} 
Ad Hoc teamwork aims to generalise the agent's cooperation capacity with novel agents \cite{stone2010ad,du2023review}
The progress in AHT has shifted from relying on preset action-selection rules to a data-driven strategy. This strategy employs methods like probabilistic models \cite{rahman2023generating}, deep learning \cite{chen2020aateam}, or reinforcement learning \cite{rahman2021towards,lou2023pecan,yan2023efficient} to predict the behaviors of various agents in different states, aiming to optimize and refine the learner's behavior.
Recent research has also explored the communication in AHT, considering scenarios where teammates follow a shared communication protocol \cite{macke2021expected, mirsky2020penny}. However, these studies are often confined to a limited multi-agent setting.
Our work stands apart from previous studies by focusing on dynamic environments based on natural language communication. We employ agents powered by LLMs to execute AHT, and explore ways to improve their performance in these settings.

\section{Conclusion and Future Work}
In this study, we investigate the AHT capabilities of LLM agents using the AvalonPlay benchmark. LLMs endow agents with powerful reasoning abilities, but challenges such as early memory forgetting and hallucination generation also arise. We address these challenges by proposing the CodeAct agent. We first employ a memory retrieval system to collect factual information from prior interactions. Subsequently, we integrate a code-driven reasoning with action to discern new teammate roles, enhancing our agent's ability to cooperate more effectively. Lastly, we employ an interpreter to execute the code and equip the agent with the self-debug capability.

Despite our initial research, there are some unresolved issues in the AHT research with language agents that require further investigation. For intance, our AvalonPlay benchmark involves moderator-guided communication for agents, making it crucial to investigate the agent's ability to autonomously decide when to communicate. Furthermore, significant work lies in reducing the effects of hallucinations and enhancing their practical application in real-world scenarios. In this study, we employ a memory retrieval system for factual information extraction, and another promising research direction involves developing systems for evaluating and verifying facts to address these challenges.

\section{Limitations}
Regarding limitations, our study does not incorporate experience pools from human players. Future research will focus on developing robust strategies for leveraging this experience and enhancing comparison with human performance.

\bibliography{custom}

\appendix
\onecolumn

\section{Global Prompt for Agents\label{appendix_prompt}}
Here, we show the global prompt used for the LLM agent in the AvalonPlay benchmark.
We provide the game rule and role assignment as the global prompt.
Figure \ref{fig:gamerule} displays the specific game rule for all agents, while Table \ref{tab:global_prompt} shows the role assignments for different agents.

\lstset{style=mystyle}
\begin{figure*}[ht!]
\begin{lstlisting}
You are playing the Avalon game with some other players. The moderator is also host, he organised this game and you need to follow his instructions correctly. Do not talk with the moderator. What the modeler said is unquestionable and absolutely true. There are six roles in the game. Merlin, Percival, and Loyal Servants of Arthur form the just side, with members only knowing their own role and voting to approve in secret ballots to make quests successful. Morgana, Assassin, and Minion belong to the evil side. They are privy to their teammates' identities and vote against in secret ballots to make the quests fail.
Tips: If you think a player is evil side, try not to let them be selected into the quest team. If this player is the leader, you can disagree with the establishment of this team. Reasoning based on facts you have observed and you cannot perceive information (such as acoustic info) other than text. Do not pretend you are other players or the moderator.
\end{lstlisting}
\caption{Game rule prompt for all LLM agents in the AvalonPlay benchmark.}
\label{fig:gamerule}
\end{figure*}

\begin{table*}[ht!]
\resizebox{1\textwidth}{!}{%
\begin{tabular}{@{}cp{1.25cm}p{17cm}@{}}
\toprule
\multicolumn{1}{l}{\textbf{}} &
  \textbf{Role} &
  \textbf{Global Prompt} \\ 
  \midrule
 &
  Merlin &
  \{\textcolor{orange}{Game Rule}\} + You are Merlin, aware of the evil presences, yet you must remain discreet in sharing this knowledge. Should the assassin identify you after the just side completes three quests, the evil side shall emerge victorious. Your objective is to ensure the success of the quests. \\
 &
  Percival &
  \{\textcolor{orange}{Game Rule}\} + As Percival, you recognize Merlin, but be wary as Morgana might mislead you by appearing as Merlin. Your allegiance is with the just side. Your objective is to ensure   the success of the quests. \\
\multirow{-3}{*}{Good} &
  Servant &
  \{\textcolor{orange}{Game Rule}\} + You are a Servant of Arthur, with no special abilities, belonging to the just side. Your objective   is to ensure the success of the quests. \\ \midrule
 &
  Morgana &
  \{\textcolor{orange}{Game Rule}\} + You are Morgana, appearing as Merlin to Percival, weaving a web of confusion. Your objective is to ensure the quest fails. \\
 &
  Assassin &
  \{\textcolor{orange}{Game Rule}\} + As the Assassin,   your critical role comes into play once the just side successfully executes   three quests; identifying Merlin can secure a win for the evil side. Your   objective is to ensure the quest fails. \\
\multirow{-3}{*}{Evil} &
  Minion &
  \{\textcolor{orange}{Game Rule}\} + You are a Minion of  Mordred, without special abilities, on the evil side. Your objective is to ensure the quest fails. \\ \bottomrule
\end{tabular}
}
  \caption{Role assignment prompt for LLM agents in various roles within the AvalonPlay benchmark.}
  \label{tab:global_prompt} 
\end{table*}

\section{An Example of Code-driven Reasoning Generation}
\label{appendix_output}

In this section, we provide a complete example of code-driven reasoning with action generation produced by the GPT-4 model, as shown in Figure \ref{list:output}.

\lstset{style=mystyle}
\begin{figure*}[t]
\begin{lstlisting}
Thought Output:
Let's start by creating a list of 7 elements to represent the players, with None as the initial value since we don't know their alignment yet. We will then apply the constraints from each round to update this list.

- From Round 2, we know that Players 2, 4, and 7 are good.
- From Round 1, we know that either Player 1 or Player 3 is bad, but not both.
- From Round 3, we know that there is at least one bad person among Player 1, Player 2, and Player 4. Since Players 2 and 4 are good, Player 1 must be bad.
- From Round 4, we know that there is at most one bad person among Player 2, Player 3, Player 5, and Player 7. Since Player 2 and Player 7 are good, and we know Player 1 is bad, Player 3 must be good, leaving Player 5 as the only possible bad player from this group.

After applying these constraints, we should be able to identify the 4 players most likely to be good.

Let's write the code to apply these constraints and find the 4 good players.

Action: Python_REPL
Action Output: 
players = [None] * 7  # Initialize the players list with None
# Apply Round 2 information
players[1] = True  # Player 2 is good
players[3] = True  # Player 4 is good
players[6] = True  # Player 7 is good

# Apply Round 1 information
# We cannot determine which one is bad yet, so we leave Player 1 and Player 3 as None

# Apply Round 3 information
# Since Player 2 and Player 4 are good, Player 1 must be bad
players[0] = False  # Player 1 is bad

# Apply Round 4 information
# Since Player 2 and Player 7 are good, and we know Player 1 is bad, Player 3 must be good
players[2] = True  # Player 3 is good
# This leaves Player 5 as the only possible bad player from this group
players[4] = False  # Player 5 is bad

# Now we can determine the remaining players' goodness based on the total number of bad players
# We already have 2 bad players (Player 1 and Player 5), so Player 6 must be good
players[5] = True  # Player 6 is good

# Print the final 4 players most likely to be good
good_players = [index + 1 for index, is_good in enumerate(players) if is_good]
print(good_players)

Observation: [2, 3, 4, 6, 7]
Thought:I now know the final answer

Final Answer: The 4 players most likely to be good are Players 2, 3, 4, and 6.
\end{lstlisting}
\caption{An example of the code-driven reasoning with action generation.}
\label{list:output}
\end{figure*}

\end{document}